\documentclass[10pt,twocolumn,letterpaper]{article}

\usepackage[pagenumbers]{cvpr} 

\usepackage[linesnumbered,vlined]{algorithm2e}
\RestyleAlgo{ruled}
\DontPrintSemicolon
\SetNoFillComment

\SetCommentSty{mycommfont}
\SetKwComment{Comment}{/* }{ */}
\SetKwInput{kwInput}{Input}
\SetKwInput{kwOutput}{Output}
\SetKwInput{kwReturn}{Return}
\SetKwInput{kwInit}{Initialize}
\SetKw{In}{in}

\newcommand{\name}{ViewExtrapolator\xspace}
\newcommand{\methodA}{guidance annealing\xspace}
\newcommand{\methodB}{resampling annealing\xspace}
\newcommand{\methodAUp}{Guidance annealing\xspace}
\newcommand{\methodBUp}{Resampling annealing\xspace}

\usepackage{colortbl}

\definecolor{cvprblue}{rgb}{0.21,0.49,0.74}
\usepackage[pagebackref,breaklinks,colorlinks,allcolors=cvprblue]{hyperref}

\def\paperID{****} 
\def\confName{CVPR}
\def\confYear{2025}

\title{Novel View Extrapolation with Video Diffusion Priors}

\author{
Kunhao Liu$^{1}$
\quad
Ling Shao$^{2}$
\quad
Shijian Lu$^{1}$
\\[2mm]
{\normalsize $^1$Nanyang Technological University\quad$^2$UCAS-Terminus AI Lab, UCAS}
}

\begin{document}

\twocolumn[{
    \renewcommand\twocolumn[1][]{#1}
    \maketitle
    \begin{center}
        \centering
        \captionsetup{type=figure}
        \includegraphics[width=\textwidth]{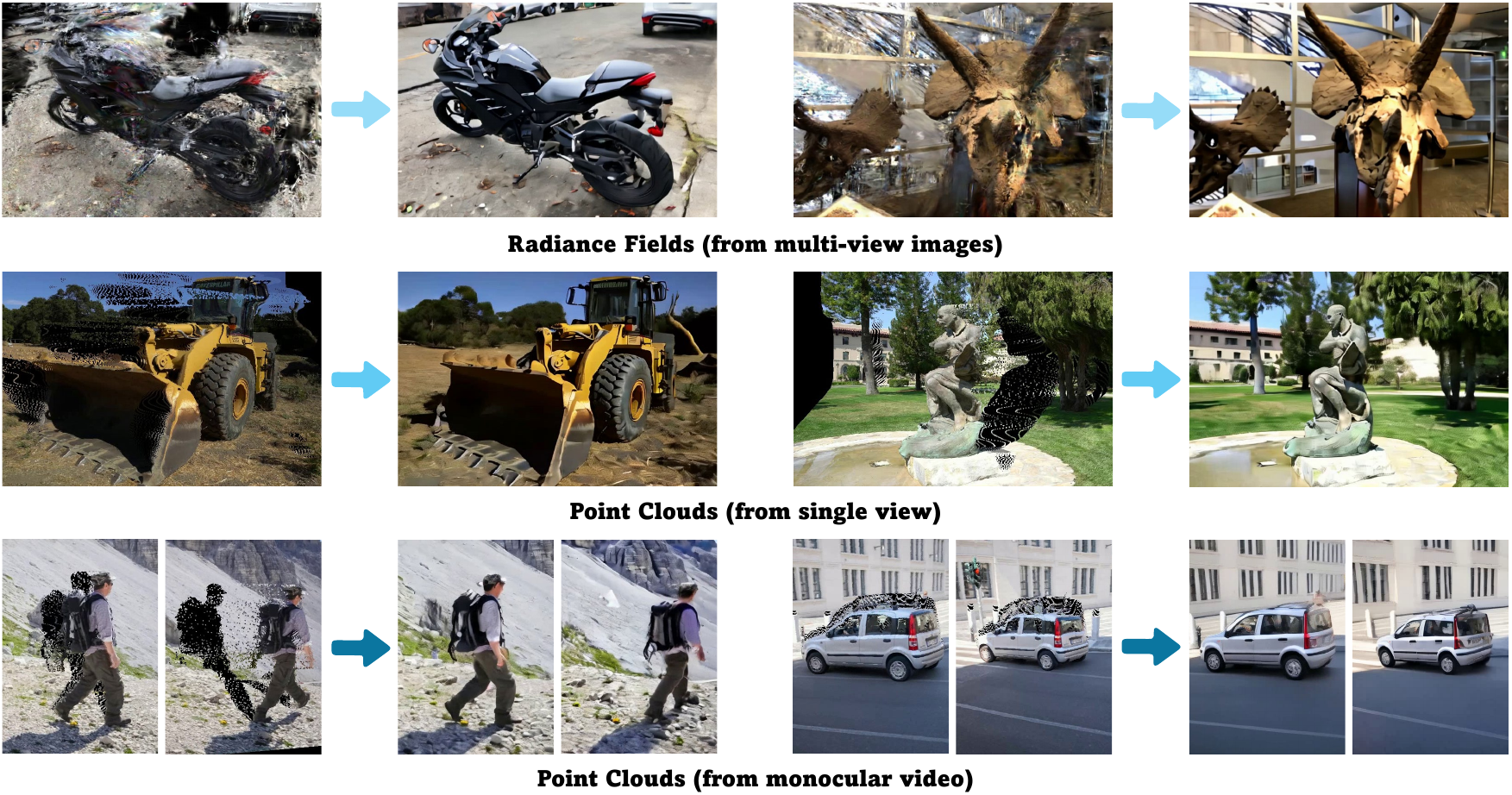}
        \captionof{figure}{We introduce \textbf{\name}, a novel approach that leverages the generative priors of Stable Video Diffusion for novel view extrapolation, where the novel views lie far beyond the range of the training views. \name effectively refines the artifact-prone renderings (left side of arrows) of radiance fields or point clouds, to more realistic renderings with fewer artifacts (right side of arrows).
        }
        \label{fig:teaser}
    \end{center}
}]

\begin{abstract}

The field of novel view synthesis has made significant strides thanks to the development of radiance field methods. However, most radiance field techniques are far better at novel view interpolation than novel view extrapolation where the synthesis novel views are far beyond the observed training views. 
We design \name, a novel view synthesis approach that leverages the generative priors of Stable Video Diffusion (SVD) for realistic novel view extrapolation. By redesigning the SVD denoising process, \name refines the artifact-prone views rendered by radiance fields, greatly enhancing the clarity and realism of the synthesized novel views. \name is a generic novel view extrapolator that can work with different types of 3D rendering such as views rendered from point clouds when only a single view or monocular video is available. Additionally, \name requires no fine-tuning of SVD, making it both data-efficient and computation-efficient. Extensive experiments demonstrate the superiority of \name in novel view extrapolation. Project page: \url{https://kunhao-liu.github.io/ViewExtrapolator/}.

\end{abstract}    
\section{Introduction}
The field of novel view synthesis has witnessed remarkable advancements, largely driven by the development of radiance field methods such as NeRF \cite{mildenhall2021nerf}, Instant-NGP \cite{muller2022instant}, 3D Gaussian Splatting \cite{kerbl20233d}, etc. These methods have revolutionized the way we render photorealistic images of novel views by learning continuous volumetric scene representations from a set of training views.

\begin{figure}[t]
    \centering
    \includegraphics[width=0.9\linewidth]{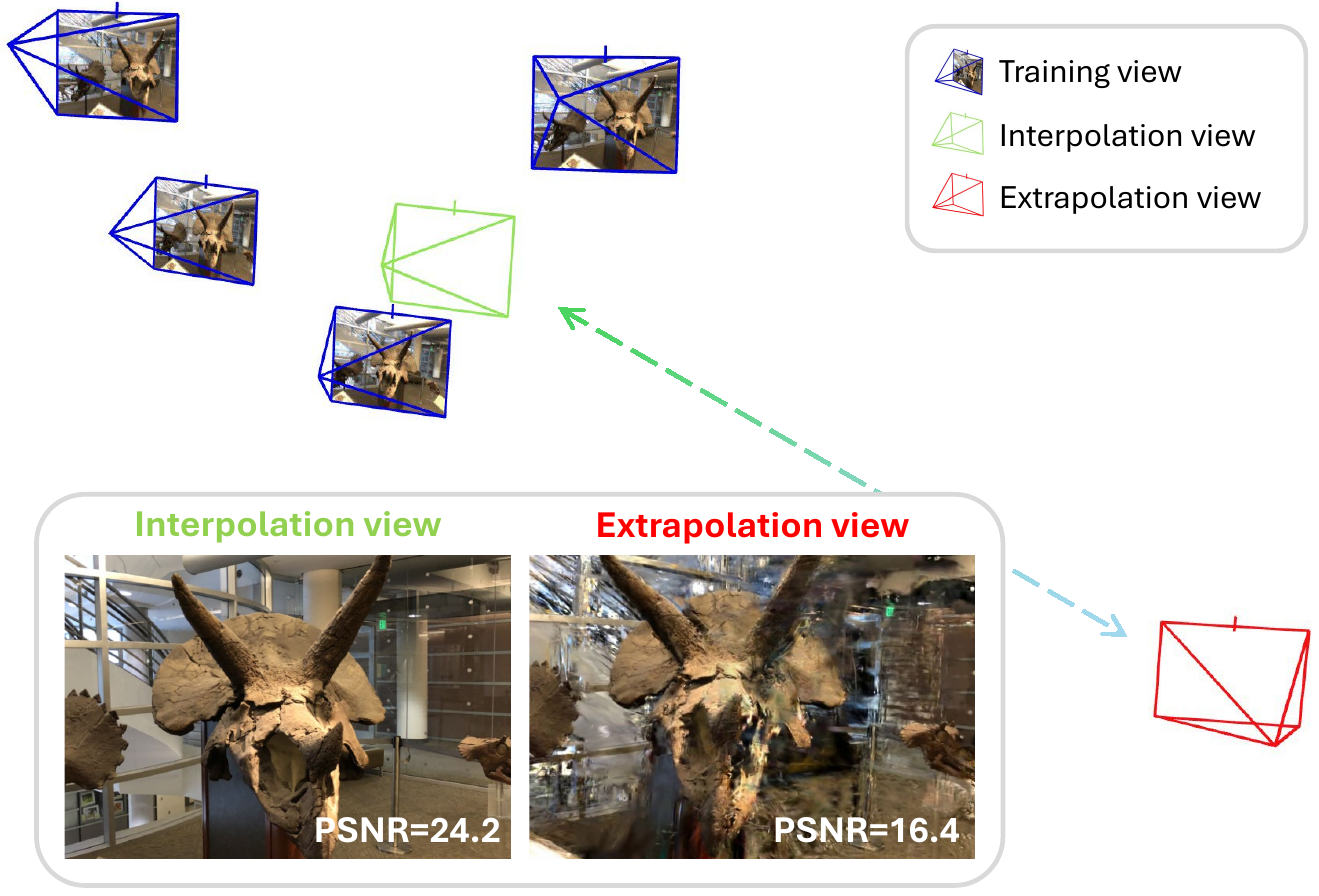}
    \caption{The setting differences between novel view \textit{interpolation} and novel view \textit{extrapolation}:  Radiance fields excel at novel view interpolation but struggle at novel view extrapolation.}
    \label{fig:setting}
\end{figure}

The success of radiance fields is especially notable in novel view \textit{interpolation} when the synthesized novel view lies within or near the convex hull of the training views. For the case of novel view \textit{extrapolation} where the novel views move significantly beyond the range of training views, most existing radiance field methods struggle due to the lack of observed training data around the novel views \cite{shih2024extranerf}. However, novel view extrapolation is crucial for delivering an immersive 3D experience, allowing users to explore reconstructed radiance fields freely beyond the initial training views.
\cref{fig:setting} illustrates the setup differences between novel view interpolation and novel view extrapolation, as well as how they affect the synthesized novel views.

We design \name, a novel view extrapolation technique that introduces the generative priors of Stable Video Diffusion (SVD) \cite{blattmann2023stable} for generating realistic extrapolative novel views. Given a reconstructed radiance field from training views with limited range, \name first renders a video that starts from a training view and gradually transits to a distant extrapolative novel view. While the early video frames exhibit high-quality renderings, artifacts gradually arise in the ensuing video frames when the view goes beyond the training views. The artifacts become especially obvious around the extrapolated regions due to the lack of observed data in training. We introduce SVD as trained over large-scale natural videos to refine the artifact-prone novel-view frames. Specifically, we redesign the denoising process to guide SVD to preserve the original scene content by modifying the ODE derivative toward the artifact-prone videos. In addition, we design \methodA and \methodB that reduce the influence of the artifacts in the denoising steps and resampling steps \cite{lugmayr2022repaint}, respectively, inpainting unseen regions and refining the visual quality throughout the denoising process effectively. 

\name has two unique features in novel view extrapolation. First, it is generic and can work with different 3D rendering approaches with little adaptation. For example, it can be directly applied to 3D renderings by point clouds as derived by depth estimation from a single view or monocular video. Second, \name is an inference-stage method that does not require fine-tuning the SVD model. This makes it both data-efficient and computation-efficient, paving the way for more applicable and accessible novel view extrapolation.

The contributions of this work can be summarized in three key aspects. \textit{First}, we introduce \name, a novel training-free pipeline that leverages the generative priors of SVD for novel view extrapolation. \textit{Second}, we design \methodA and \methodB that eliminate artifacts and enable high-quality inpainting of unseen regions, enhancing the visual fidelity of the rendered novel views effectively. \textit{Third}, extensive experiments over various 3D rendering approaches demonstrate the superiority and broad applicability of \name in novel view extrapolation.

\section{Related Work}
\paragraph{Radiance fields.}
Radiance fields \cite{mildenhall2021nerf} have emerged as a powerful representation of 3D scenes, driving advancements in novel view synthesis. They model 3D space by mapping radiance and density to arbitrary 3D coordinates, where pixel colors are rendered by aggregating the radiance values of sampled 3D points through volume rendering \cite{max1995optical}. Radiance fields can be implemented using various methods, including MLPs \cite{mildenhall2021nerf, barron2021mip, barron2022mip, zhang2020nerf++}, decomposed tensors \cite{chen2022tensorf, chan2022efficient, fridovich2023k, liu2023stylerf, liu2023weakly}, hash tables \cite{muller2022instant}, voxels \cite{sun2022direct, fridovich2022plenoxels}, and 3D Gaussians \cite{kerbl20233d, lu2024scaffold, liu2024stylegaussian}. Numerous studies have been proposed to enhance the view synthesis process. For instance, Mip-NeRF \cite{barron2021mip, barron2022mip} improves rendering quality using anti-aliased conical frustums. Instant-NGP \cite{muller2022instant} accelerates training speed by modeling 3D volumes with multi-resolution hash tables. 3D Gaussian Splatting \cite{kerbl20233d} achieves real-time rendering through rasterization with explicitly parameterized 3D Gaussians. However, these approaches generally require dense scene observations and lack the generative capacity for extrapolating beyond observed views, limiting their effectiveness in novel view extrapolation. While methods like ExtraNeRF \cite{shih2024extranerf} and RapNeRF \cite{zhang2022ray} attempt to address novel view extrapolation, ExtraNeRF's extrapolation range is limited, and RapNeRF is restricted to object-level view synthesis. In contrast, \name can render scene-level realistic novel views that lie far beyond the range of the training views.

\paragraph{Diffusion priors for view synthesis.}
Recent work has explored the generative priors of diffusion models \cite{ho2020denoising} for novel view synthesis. Early efforts focused on distilling the knowledge of 2D text-to-image diffusion models \cite{rombach2022high} into 3D using Score Distillation Sampling \cite{poole2022dreamfusion, wang2023score}, synthesizing 3D objects from text and images \cite{huang2023dreamtime, wang2024prolificdreamer, lin2023magic3d, tang2023dreamgaussian}. Several studies fine-tune or train 2D diffusion models on multi-view or camera-pose-conditioned datasets to strengthen 3D priors \cite{liu2023zero, wang2023imagedream, shi2023mvdream, shi2023zero123++, liu2023syncdreamer, hollein2024viewdiff, wu2024reconfusion, liu2023zero, watson2022novel, gu2023nerfdiff, chan2023generative, sargent2023zeronvs}, though most of them focus on object-level synthesis. For scene-level synthesis, approaches like ExtraNeRF \cite{shih2024extranerf}, DiffusioNeRF \cite{wynn2023diffusionerf}, and Nerfbusters \cite{warburg2023nerfbusters} incorporate geometry-informed diffusion models for improved scene-level 3D reconstruction, while methods like Zero-NVS \cite{sargent2023zeronvs}, Reconfusion \cite{wu2024reconfusion}, and CAT3D \cite{gao2024cat3d} employ diffusion models trained on large-scale multi-view datasets to enable scene-level few-shot reconstruction. In addition, MotionCtrl \cite{wang2024motionctrl}, CameraCtrl \cite{he2024cameractrl}, ViVid-1-to-3 \cite{kwak2024vivid}, and SV3D \cite{voleti2025sv3d} leverage video diffusion models fine-tuned on camera trajectories for view synthesis, whereas NVS-solver \cite{you2024nvs} and CamTrol \cite{hou2024training} utilize a training-free approach for camera control. Different from these developments, we propose a training-free approach for novel view extrapolation with video diffusion priors, paving a more applicable and accessible way in novel view synthesis.
\section{Method}

\begin{figure}[t]
    \centering
    \includegraphics[width=\linewidth]{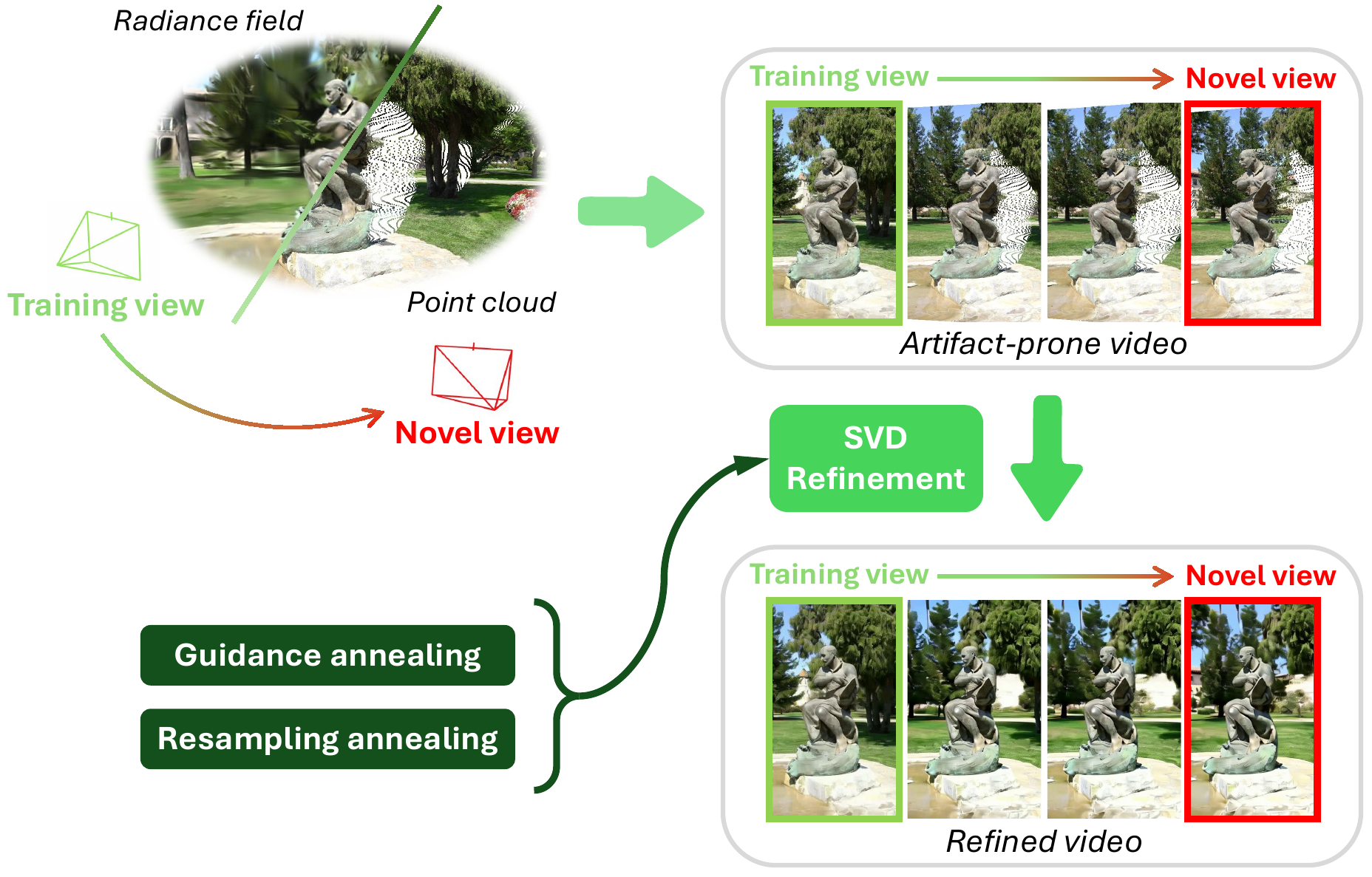}
    \caption{\textbf{Overview of the proposed \name.} We render an artifact-prone video from the closest training view to an extrapolative novel view, and then refine it by guiding SVD to preserve the original scene content and eliminate the artifacts with \methodA and \methodB.}
    \label{fig:overview}
\end{figure}

We tackle the challenges of novel view extrapolation by leveraging the generative priors of a large-scale video diffusion model SVD (\cref{sec:SVD}) for refining artifact-prone videos as rendered by radiance fields or point clouds (\cref{sec:render}). Specifically, we guide the SVD model to preserve the original scene content by modifying the ODE derivative towards the artifact-prone videos (\cref{sec:guidance}). Additionally, we design \methodA and \methodB, which enable SVD to effectively refine the artifact-prone videos during the denoising process (\cref{sec:refine}). \cref{fig:overview} illustrates the overview of the proposed \name.

\subsection{Preliminaries on Stable Video Diffusion}
\label{sec:SVD}
SVD \cite{blattmann2023stable} is an image-to-video diffusion model that conditions on an input image. By default, it generates a natural video that starts with the conditional image and autonomously evolves with camera movements and scene dynamics. As a diffusion model \cite{ho2020denoising}, SVD produces the video by progressively denoising a Gaussian noise.  Given the noisy video latent $\mathbf{x}_t$ and the noise level $\sigma_t$ at the diffusion time step $t \in [1,T]$, SVD parameterizes the denoising process following the EDM pre-conditioning framework \cite{karras2022elucidating}:
\begin{equation}
\label{eq:EDM}
    \hat{\mathbf{x}}_0 = c_{\mathrm{skip}}(\sigma_t)\mathbf{x}_t + c_{\mathrm{out}}(\sigma_t)F_{\boldsymbol{\theta}}(c_{\mathrm{in}}(\sigma_t)\mathbf{x}_t; c_{\mathrm{noise}}(\sigma_t)),
\end{equation}
where $\hat{\mathbf{x}}_0$ is the predicted clean video at the current time step $t$, $c_{\mathrm{skip}}, c_{\mathrm{out}}, c_{\mathrm{in}},$ and $c_{\mathrm{noise}}$ denote the predefined preconditioning functions, and$F_{\boldsymbol{\theta}}$ is the trainable network with parameters $\boldsymbol {\theta}$. With the current predicted clean video $\hat{\mathbf{x}}_0$, the ODE derivative can be computed by:
\begin{equation}
\label{eq:derivative}
    \mathrm{d}\mathbf{x} = (\mathbf{x}_t - \hat{\mathbf{x}}_0) / \sigma_t.
\end{equation}
We can then obtain the estimated denoised sample $\mathbf{x}_{t-1}$ at the previous time step by:
\begin{equation}
\label{eq:denoise_detail}
    \mathbf{x}_{t-1} = \mathbf{x}_{t} + \mathrm{d}\mathbf{x}(\sigma_{t-1}-\sigma_t).
\end{equation}
The above denoising process can be abstracted into two steps: 1) Predicting the clean video given the current noisy latent: $\mathrm{Predict}(\mathbf{x}_t) = \hat{\mathbf{x}}_0$ as defined in \cref{eq:EDM}; 2) Denoising the current latent to get the previous-time-step latent: $\mathrm{Denoise}(\mathbf{x}_t, \hat{\mathbf{x}}_0) = \mathbf{x}_{t-1}$ as defined in \cref{eq:derivative,eq:denoise_detail}. By repeating the two steps, SVD progressively denoises the latent and finally produces a clean video $\mathbf{x}_0$.

\subsection{Rendering Artifact-prone Videos}
\label{sec:render}
Given multiple training views and an extrapolative novel view lying far from the training views, a radiance field can be trained with techniques like 3D Gaussian Splatting \cite{kerbl20233d} and a video can be further rendered that starts from the nearest training view and gradually transitions to the extrapolative novel view. When only a single view or monocular video is available, depth can be estimated by using off-the-shelf image or video depth estimators such as UniDepth \cite{piccinelli2024unidepth} or DepthCrafter \cite{hu2024depthcrafter}. With the estimated depth, the image or monocular video can be projected into a point cloud for rendering a video starting from the initial view to the extrapolative novel view.

The initial video frames usually exhibit a clean and accurate appearance since the rendered video starts from one observed training view. However, significant artifacts and unnatural looking appear as the view of the rendered video frames extends beyond the range of the training views. Nevertheless, the rendered videos still retain valuable information about the scene’s geometry and appearance. Given that SVD is trained with large-scale natural videos, we exploit the distribution of natural videos in SVD to inpaint and refine the rendered artifact-prone videos.

\begin{algorithm}[t]
\caption{Video refinement with guidance annealing and resampling annealing.}\label{alg:refine}
\kwInput{ artifact-prone video $\tilde{\mathbf{x}}$, opacity mask $\mathbf{m}$}

$x_T \sim \mathcal{N}(\mathbf{0}, \mathbf{1})$\;
\For{$t=T, \dotsc, 1$}{

    \eIf{$t > T - T^\mathrm{guide}$}{ 
      \For{$r=1, \dotsc, R$}{
        $\hat{\mathbf{x}}_0 = \mathrm{Predict}(\mathbf{x}_t)$\;
        \eIf{$r \leq R^\mathrm{guide}$}{
            $\hat{\mathbf{x}}_0^\mathrm{dir} = \tilde{\mathbf{x}} \odot \mathbf{m} + \hat{\mathbf{x}}_0 \odot (1-\mathbf{m})$\;
        }{
            $\hat{\mathbf{x}}_0^\mathrm{dir} = \hat{\mathbf{x}}_0$\;
        }
        $\mathbf{x}_{t-1} = \mathrm{Denoise}(\mathbf{x}_t, \hat{\mathbf{x}}_0^\mathrm{dir})$\;
        \If{$r < R$}{
            $\mathbf{x}_t \sim \mathcal{N}(\hat{\mathbf{x}}_0^\mathrm{dir}, \sigma_t)$
        }
      }
    }{
    $\hat{\mathbf{x}}_0 = \mathrm{Predict}(\mathbf{x}_t)$\;
    $\mathbf{x}_{t-1} = \mathrm{Denoise}(\mathbf{x}_t, \hat{\mathbf{x}}_0)$
    }
}
\textbf{return} $\mathbf{x}_0$
\end{algorithm}

\subsection{Guidance with Input Videos}
\label{sec:guidance}
Given the rendered artifact-prone video  $\tilde{\mathbf{x}}$, our goal is to refine it for a more natural appearance, reducing artifacts while preserving the original content. Since the first frame of  $\tilde{\mathbf{x}}$ contains minimal artifacts, it can effectively serve as the image condition for SVD. Beyond the image condition, we also need to condition SVD on the remainder of the video to ensure that the output video retains the original content, including camera movement, scene dynamics, and geometry. We can interpret \cref{eq:derivative} as denoising the noisy latent at each time step towards the direction of the predicted clean video $\hat{\mathbf{x}}_0$. To guide the denoising process towards $\tilde{\mathbf{x}}$, we can replace the $\hat{\mathbf{x}}_0$ in \cref{eq:derivative} with $\tilde{\mathbf{x}}$. However, since $\tilde{\mathbf{x}}$ may contain regions of the scene that are not fully captured, we also need to leverage SVD for multi-view consistent video inpainting. This can be achieved by allowing SVD to denoise the unseen parts without the guidance from $\tilde{\mathbf{x}}$. Given the opacity mask $\mathbf{m}$ indicating the unseen parts, we can obtain the denoising direction as:
\begin{equation}
    \hat{\mathbf{x}}_0^\mathrm{dir} = \tilde{\mathbf{x}} \odot \mathbf{m} + \hat{\mathbf{x}}_0 \odot (1-\mathbf{m}),
\end{equation}
where the seen parts are used to guide the denoising process, and the unseen parts are inpainted by SVD. Then we can replace the denoising direction in the original denoising step to achieve guided denoising:
\begin{equation}
\label{eq:guided_denoise}
\mathbf{x}_{t-1} = \mathrm{Denoise}(\mathbf{x}_t, \hat{\mathbf{x}}_0^\mathrm{dir}).
\end{equation}

\begin{table}[t]
    \definecolor{red}{rgb}{1,0.6,0.6}
    \definecolor{orange}{rgb}{1,0.8,0.6}
    \definecolor{yellow}{rgb}{1,1,0.6}
    \centering
    \resizebox{\linewidth}{!}{
    \begin{tabular}{l|ccc}
        \toprule
        \textbf{Methods} & \textbf{SSIM $\uparrow$} & \textbf{PSNR $\uparrow$} & \textbf{LPIPS $\downarrow$} \\
        \midrule
        3DGS        & \cellcolor{yellow} 0.416 & 14.46 & \cellcolor{yellow} 0.429 \\
        DRGS     & 0.406 & \cellcolor{yellow} 14.68 & 0.457 \\
        \textbf{\name (video)} & \cellcolor{orange} 0.427 & \cellcolor{orange} 14.83 & \cellcolor{orange} 0.379 \\
        \textbf{\name (3DGS)} & \cellcolor{red} 0.460 & \cellcolor{red} 15.46 & \cellcolor{red} 0.378 \\
        \midrule
        \name w/o GA &  0.442 &  15.14 &  0.448 \\
        \name w/o RA &  0.456 &  15.33 &  0.382 \\
        
        \bottomrule
    \end{tabular}
    }
    \caption{\textbf{Quantitative comparisons and ablation studies.} The first four rows present the comparison results, while the last two rows show the ablation studies. 
    \name w/o GA denotes results without \methodA, and \name w/o RA denotes results without \methodB.}
    \label{tab:comparison}
\end{table}

\begin{figure*}[t]
    \centering
    \includegraphics[width=\linewidth]{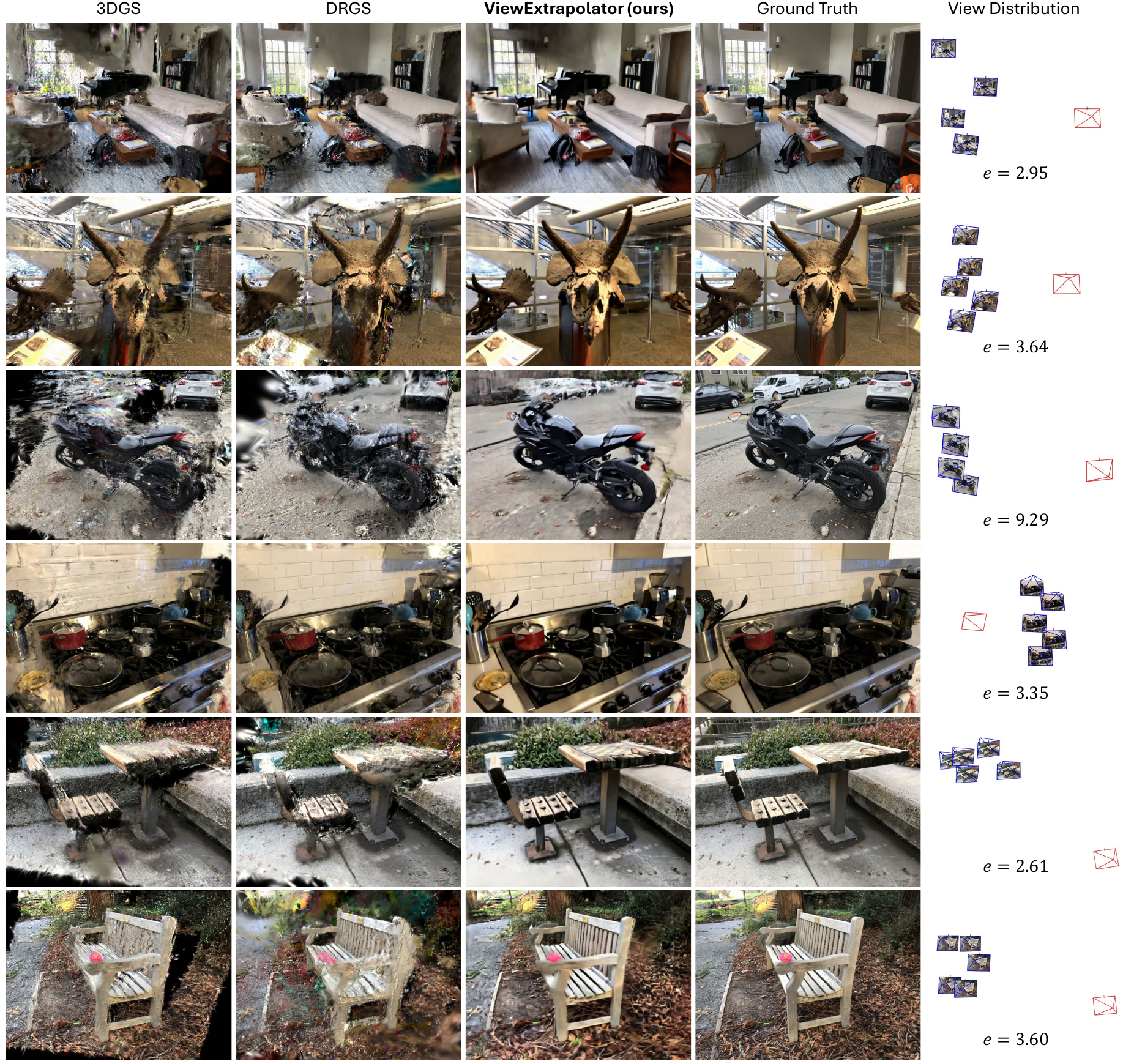}
    \caption{\textbf{Qualitative comparisons.} We compare \name with 3DGS and DRGS on novel view extrapolation. \name demonstrates superior generation quality with much fewer artifacts. The last column shows the distribution of training and test views as well as the corresponding extrapolation degree $e$. Zoom in for details.}
    \label{fig:comparison}
\end{figure*}

\subsection{Video Refinement}
\label{sec:refine}

\paragraph{\methodAUp.} While the denoising process in  \cref{eq:guided_denoise} is guided by the artifact-prone video $\tilde{\mathbf{x}}$, it alone cannot remove the artifacts within $\tilde{\mathbf{x}}$ which predominantly exist in the finer details of the video. Since the diffusion models gradually add details during the denoising process, we guide the denoising process in \cref{eq:guided_denoise} during the first $T^\mathrm{guide}$ denoising steps only, as indicated in line 3 of \cref{alg:refine}. During the rest unguided steps of the denoising process, SVD remains conditioned on the first frame of $\tilde{\mathbf{x}}$ and continues denoising the latent produced after $T^\mathrm{guide}$ guided steps. This approach allows SVD to generate natural video details based on the clean first frame while retaining the coarse structure from the previously denoised latent, thus reducing the artifacts contained in $\tilde{\mathbf{x}}$ and generating more natural and consistent details.

\paragraph{\methodBUp.} However, artifacts in the latent accumulate during the first $T^\mathrm{guide}$ denoising steps (as each guided step with $\tilde{\mathbf{x}}$ introduces artifacts), which could become too dominant for SVD to refine in the subsequent unguided denoising steps. Therefore, it is necessary for SVD to refine the denoised latent throughout the initial $T^\mathrm{guide}$ denoising steps as well. Drawing inspiration from the resampling technique \cite{lugmayr2022repaint} that reduces artifacts by repeating a denoising step multiple times, we incorporate $R$ resampling steps at each of the $T^\mathrm{guide}$ denoising steps, as indicated in line 4 of \cref{alg:refine}. Specifically, during each guided denoising step $t$, after obtaining the denoised latent $\mathbf{x}_{t-1}$ from the previous time steps with \cref{eq:guided_denoise}, we diffuse $\mathbf{x}_{t-1}$ back to $\mathbf{x}_t$ as $\mathbf{x}_t \sim \mathcal{N}(\hat{\mathbf{x}}_0^\mathrm{dir}, \sigma_t)$. This is followed by another round of denoising over $\mathbf{x}_t$ as defined in \cref{eq:guided_denoise}.
However, the resampling technique in \cite{lugmayr2022repaint} is originally designed for inpainting tasks where the goal is to preserve the visible regions unaltered, whereas we need to refine artifacts in the visible regions. Since SVD can denoise the latent towards the direction of a natural video that contains few artifacts, we apply the guidance in \cref{eq:guided_denoise} only for the first $R^{\mathrm{guide}}$ resampling steps in each denoising step, allowing SVD to denoise without the guidance of $\tilde{\mathbf{x}}$ in the remaining resampling steps, as indicated in line 6 of \cref{alg:refine}. During these unguided resampling steps, SVD denoises the latent towards a more natural video, effectively reducing the artifacts introduced in the guided steps. 

The above \methodA and \methodB can be combined and formulated as:
\begin{equation}
    \hat{\mathbf{x}}_0^\mathrm{dir} = \begin{cases} 
        \hat{\mathbf{x}}_0, \quad  \text{if } t \le T-T^\mathrm{guide} \text{ and } r > R^\mathrm{guide} \\
        \tilde{\mathbf{x}} \odot \mathbf{m} + \hat{\mathbf{x}}_0 \odot (1-\mathbf{m}), \quad  \text{else}  
        \end{cases},
\end{equation}
where $t\in[1,T]$ is the denoising time step and $r\in[1,R]$ is the resampling step.
With the guidance from the artifact-prone video and the video refinement with \methodA and \methodB, we derive the complete denoising algorithm as illustrated in \cref{alg:refine}.

\section{Experiments}

We conduct extensive experiments to evaluate the proposed \name on novel view extrapolation. For 3D renderings from radiance fields, we describe the settings of the evaluation dataset in detail (\cref{sec:dataset}) and benchmark \name with existing methods both qualitatively and quantitatively (\cref{sec:comparison}). For 3D renderings with point clouds, since novel view synthesis from a single view and monocular video is inherently under-constrained, we focus on qualitative evaluations only for highlighting the broad applicability of our method (\cref{sec:different}). In addition, we conduct ablation studies to validate the necessity and effectiveness of our key design choices (\cref{sec:ablation}). The implementation details are provided in the appendix.

\subsection{Dataset}
\label{sec:dataset}
\begin{figure}[t]
    \centering
    \includegraphics[width=\linewidth]{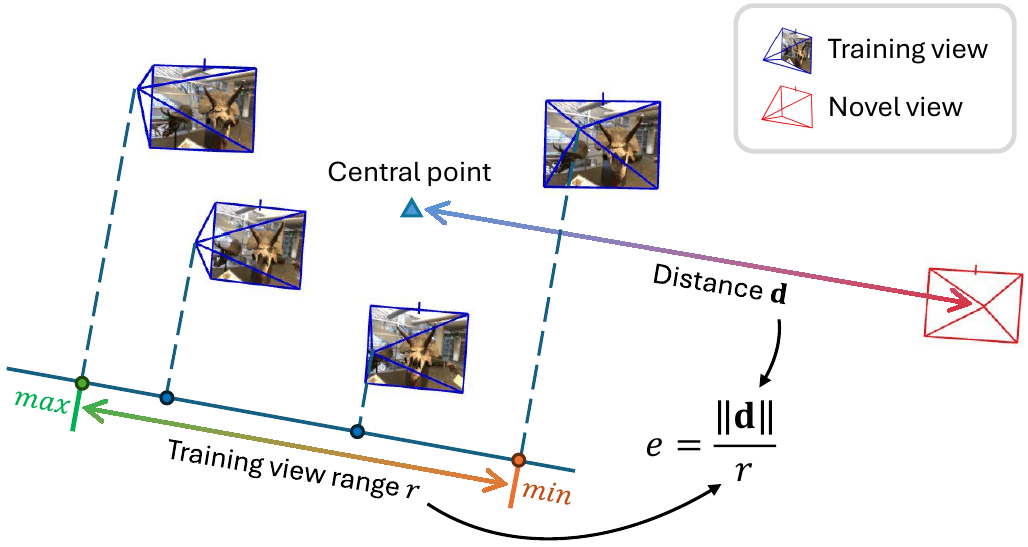}
    \caption{The definition of \textbf{extrapolation degree} $e$ by the ratio between $\mathbf{d}$ and $r$ ($\mathbf{d}$ stands for the distance between the novel view and the central point of training views, and $r$ stands for the training view range as the maximum extent of the training views along the direction of $\mathbf{d}$). A higher $e$ means that the novel view is farther away from the training views.}
    \label{fig:dataset}
\end{figure}

Effective evaluation of novel view extrapolation requires a dataset where the test views lie significantly beyond the training views for each scene. To create such a dataset, it is crucial to define a metric that can quantify and measure the distance of a novel view from a set of training views. This metric should increase as the novel view moves further away from the training views. In addition, it should be invariant to the scene scale, as camera poses of real-world data are often scaled arbitrarily \cite{schoenberger2016sfm}. To this end, we formulate an intuitive metric called extrapolation degree $e$ as illustrated in \cref{fig:dataset}.
Given a set of training views \( \mathbf{P} = \{ \mathbf{p}_1, \mathbf{p}_2, \dots, \mathbf{p}_N \} \) and a test novel view \( \mathbf{q} \) with similar viewing directions, the distance \( \mathbf{d} \) from $\mathbf{q}$ to the centroid of $\mathbf{P}$ can be computed by: $\mathbf{d} = \frac{1}{N} \sum_{i=1}^{N} \mathbf{p}_i - \mathbf{q}$. Another parameter \( r \) measuring the range of $\mathbf{P}$ can be derived by the maximum extent of $\mathbf{P}$ along the direction of \( \mathbf{d} \) as follows:
\begin{equation}
   r = \max_i ( \mathbf{p}_i \cdot \frac{\mathbf{d}}{\|\mathbf{d}\|} ) - \min_i ( \mathbf{p}_i \cdot \frac{\mathbf{d}}{\|\mathbf{d}\|} ).
\end{equation}
The extrapolation degree \( e \) can thus be defined by:
\begin{equation}
e = \frac{\|\mathbf{d}\|}{r}.
\end{equation} The defined extrapolation degree $e$ thus increases proportionally with $\|\mathbf{d}\| $ when the novel view moves further away from the training views and inversely with \( r \) when the training views have more extensive coverage of the scene. It also ensures that the novel view lies outside the convex hull of the training views when $e > 1$. 
Thus, a novel view with $e > 1$ will likely be in the novel view extrapolation setting.

\begin{figure}[t]
    \centering
    \includegraphics[width=\linewidth]{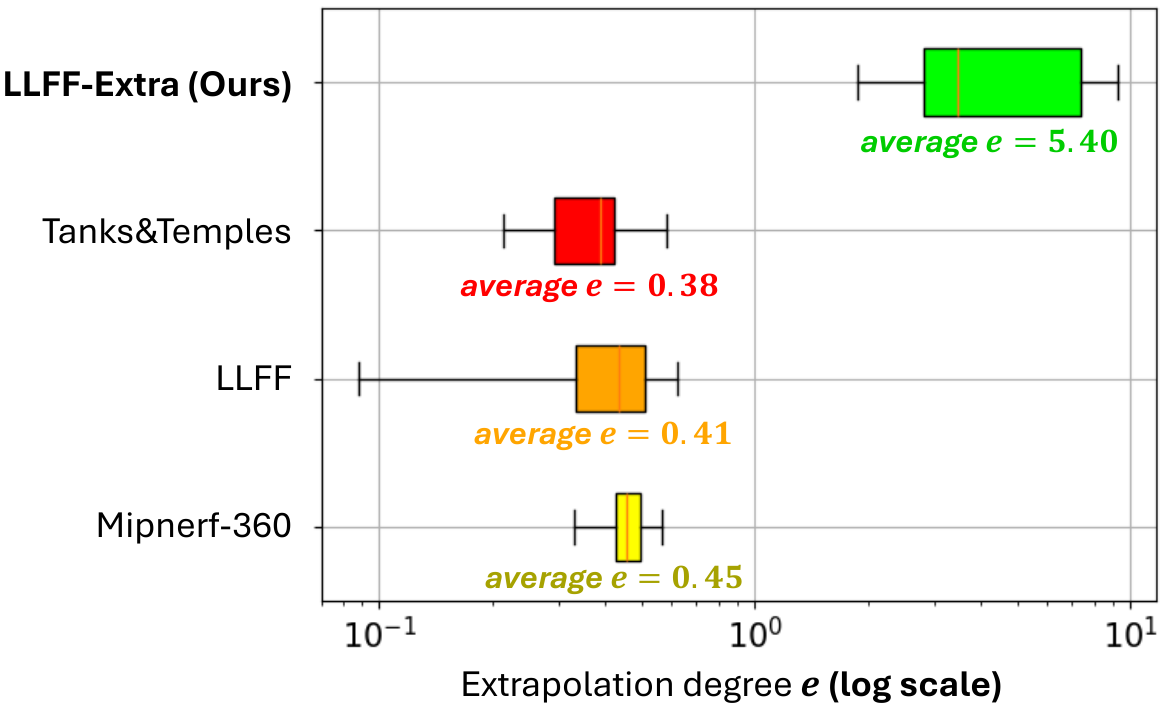}
    \caption{\textbf{Distributions of extrapolation degree $e$} across existing benchmarks and our proposed LLFF-Extra.  Unlike LLFF-Extra, all existing benchmarks exhibit a small $e$, indicating that they predominantly focus on the evaluation of novel view interpolation instead of extrapolation.}
    \label{fig:difference}
\end{figure}

Most existing benchmarks such as LLFF \cite{mildenhall2019local} and Mipnerf-360 \cite{barron2022mip} are not suitable for evaluating novel view extrapolation as they take an interpolation setting (with small $e$) by default as illustrated in \cref{fig:difference}. We construct LLFF-Extra, a new benchmark that has large $e$ and can be straightly employed to evaluate novel view extrapolation. Specifically, we use 12 scenes from LLFF and select the training views and test novel views with $e=5.4$ on average, leading to the first benchmark that can be adopted in the future study of novel view extrapolation.

\subsection{Benchmarking}
\label{sec:comparison}

\begin{figure*}
    \centering
    \includegraphics[width=\linewidth]{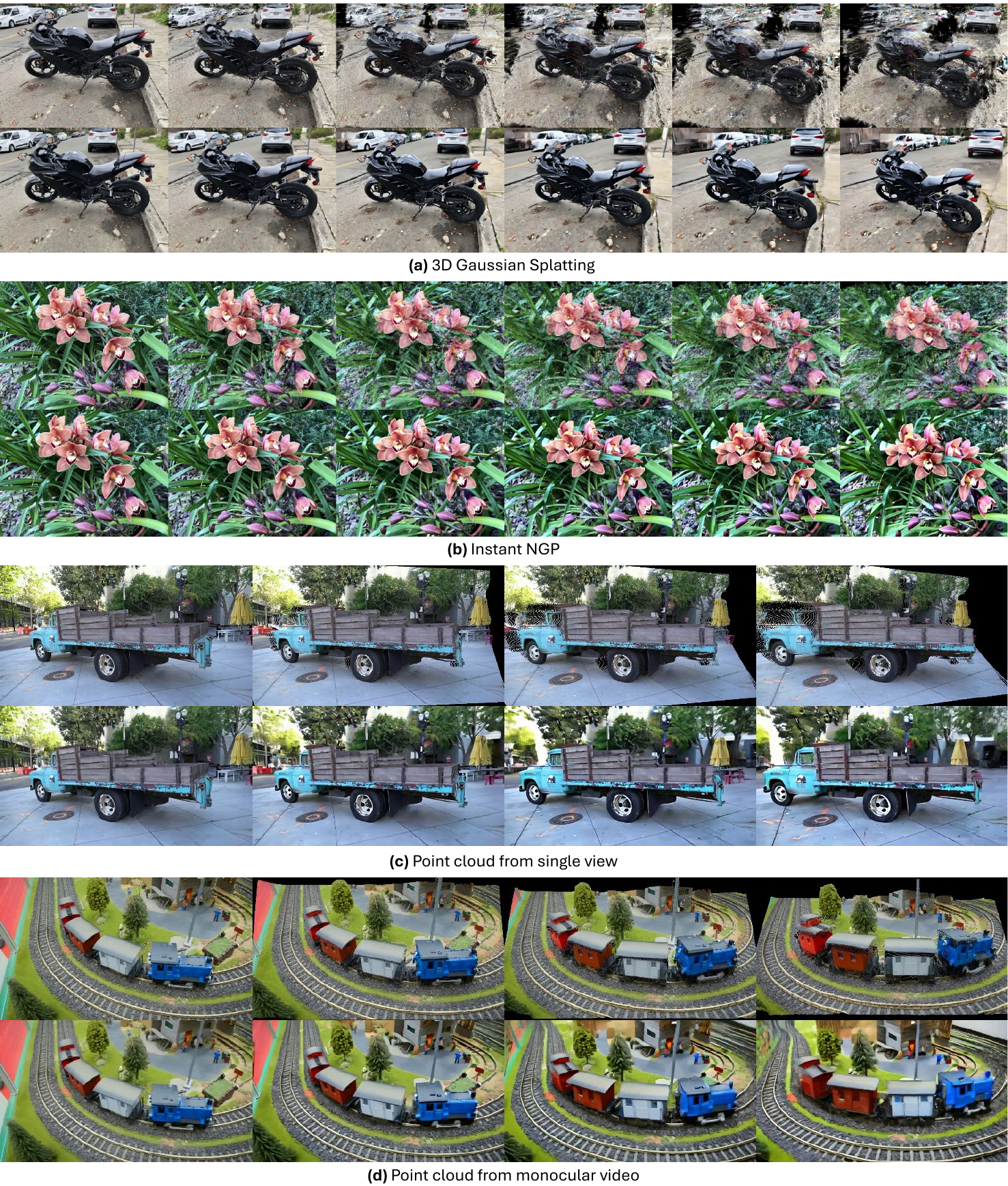}
    \caption{\textbf{Results from different rendering methods.} Our method can refine view sequences rendered from \textbf{(a)} 3D Gaussian Splatting, \textbf{(b)} Instant-NGP, and point cloud from \textbf{(c)} a single view or \textbf{(d)} monocular video. (The top row in each section is the rendered artifact-prone video and the bottom row is the refined video.)}
    \label{fig:results}
\end{figure*}

\begin{figure*}[t]
    \centering
    \includegraphics[width=\linewidth]{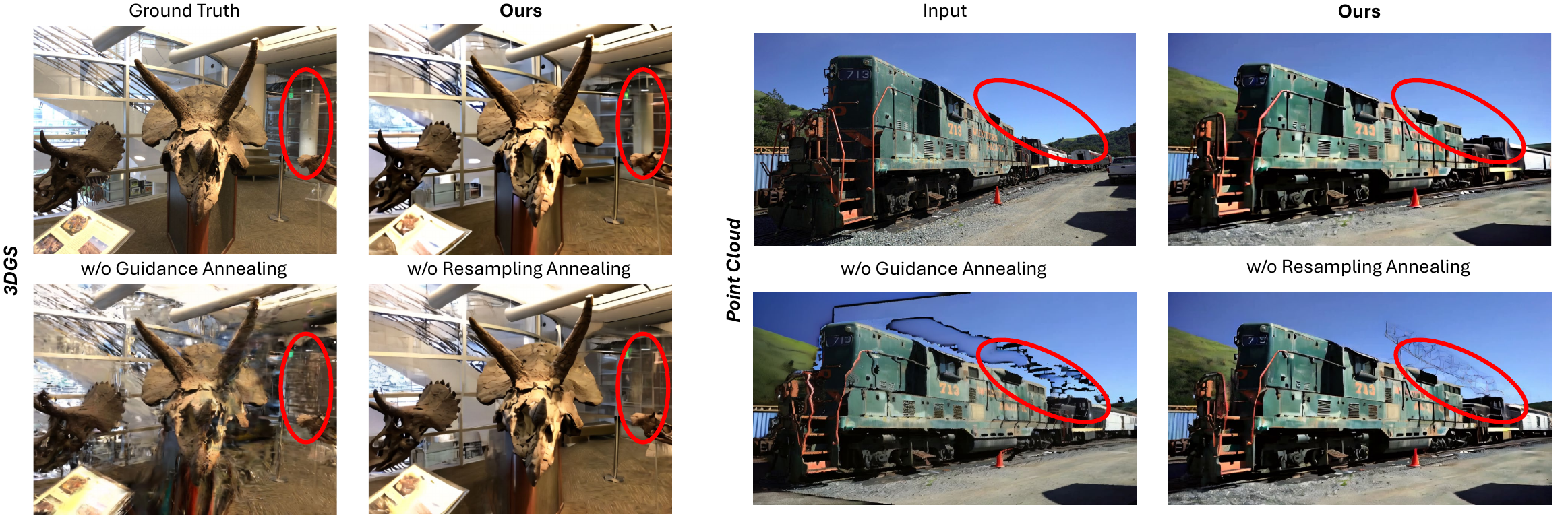}
    \caption{\textbf{Ablation studies.} We show the ablation results for 3DGS and point cloud renderings. As point clouds are used for single-image novel view extrapolation without ground truth, we show the input image for reference instead.
    As highlighted in the \textcolor{red}{red circles}, both \methodA and \methodB are essential for artifact refinement. Please zoom in for details.}
    \label{fig:ablation}
\end{figure*}

We benchmark \name with the original 3D Gaussian Splatting (3DGS) \cite{kerbl20233d} and its depth-regularized variant DRGS~\cite{chung2024depth} which incorporates depth~\cite{bhat2023zoedepth} as a geometric prior to enhance the reconstruction quality. By using 3DGS renderings as the artifact-prone videos, we employ the refined video frames (Ours (video) in \cref{tab:comparison}) to tune the pre-trained 3DGS model and evaluate renderings from the tuned 3DGS model (Ours (3DGS) in \cref{tab:comparison}) for fair comparison. The quantitative evaluations involve standard novel view synthesis metrics including SSIM, PSNR, and LPIPS \cite{zhang2018unreasonable}. We would highlight that LPIPS is more suitable for evaluating novel view extrapolation which is more toward a generative instead of regressive task with many unseen parts to generate in extrapolative views.

\name surpasses 3DGS and DRGS both qualitatively and quantitively, achieving superior visual reconstruction with much fewer artifacts as illustrated in \cref{fig:comparison} and \cref{tab:comparison}. One key observation is that 3DGS renderings degrade severely under the novel view extrapolation setting. Additionally, the incorporation of depth priors in DRGS does not lead to much improvement. Both experiments underscore that the core challenge in novel view extrapolation lies with the lack of observations in extrapolated views and direct incorporation of geometry priors alone will not solve the problem. As a comparison, \name achieves substantial improvement in perceptual quality (LPIPS), demonstrating the effectiveness of novel view refinement with generative priors from SVD.

\subsection{Broad Applicability}
\label{sec:different}

The proposed \name is versatile and can generalize to various 3D rendering approaches that often come with different types of artifacts in novel view extrapolation. We verify this feature over renderings by radiance fields and point clouds. For radiance fields, we test \name over Instant-NGP~\cite{muller2022instant}. Unlike 3DGS artifacts with noisy clusters of 3D Gaussians, Instant-NGP often produces blurry and fine-grained artifacts. \name corrects both types of artifacts effectively as illustrated in \cref{fig:results} (a, b). For point clouds, we evaluate \name over point-cloud renderings when only a single view or monocular video is available. As \cref{fig:results} (c,d) shows, \name removes the unique point artifacts effectively. The above studies demonstrate the superior generalization and flexibility of \name, highlighting its broad applicability across various scenarios with little tuning.

\subsection{Ablation Studies}
\label{sec:ablation}

We conduct ablation studies to examine how the proposed \methodA and \methodB contribute to novel view extrapolation. In the studies, we apply guidance at every diffusion time step and resampling step, respectively, for verifying \methodA and \methodB. As \cref{fig:ablation} and \cref{tab:comparison} show, only partial artifacts are refined without \methodB while most artifacts remain intact without \methodA. This verifies the crucial role of artifact refinement with \methodA and \methodB.

\section{Conclusion}
We present \name, a novel and training-free approach for novel view extrapolation. While current radiance field methods struggle to synthesize novel views that lie far beyond the range of the training views, \name is able to render realistic views by leveraging the generative priors of SVD. We refine the artifact-prone views rendered by radiance fields by guiding SVD to preserve the scene content and eliminate the artifacts at the same time. \name demonstrates superior novel view extrapolation quality compared to current methods and can also be applied to point cloud renderings when only a single view or monocular video is available.

{
    \small
    \bibliographystyle{ieeenat_fullname}
    \bibliography{main}
}

\clearpage

\appendix

\twocolumn[{
    \renewcommand\twocolumn[1][]{#1}
    \maketitlesupplementary
    \begin{center}
        \centering
        \captionsetup{type=figure}
        \includegraphics[width=\textwidth]{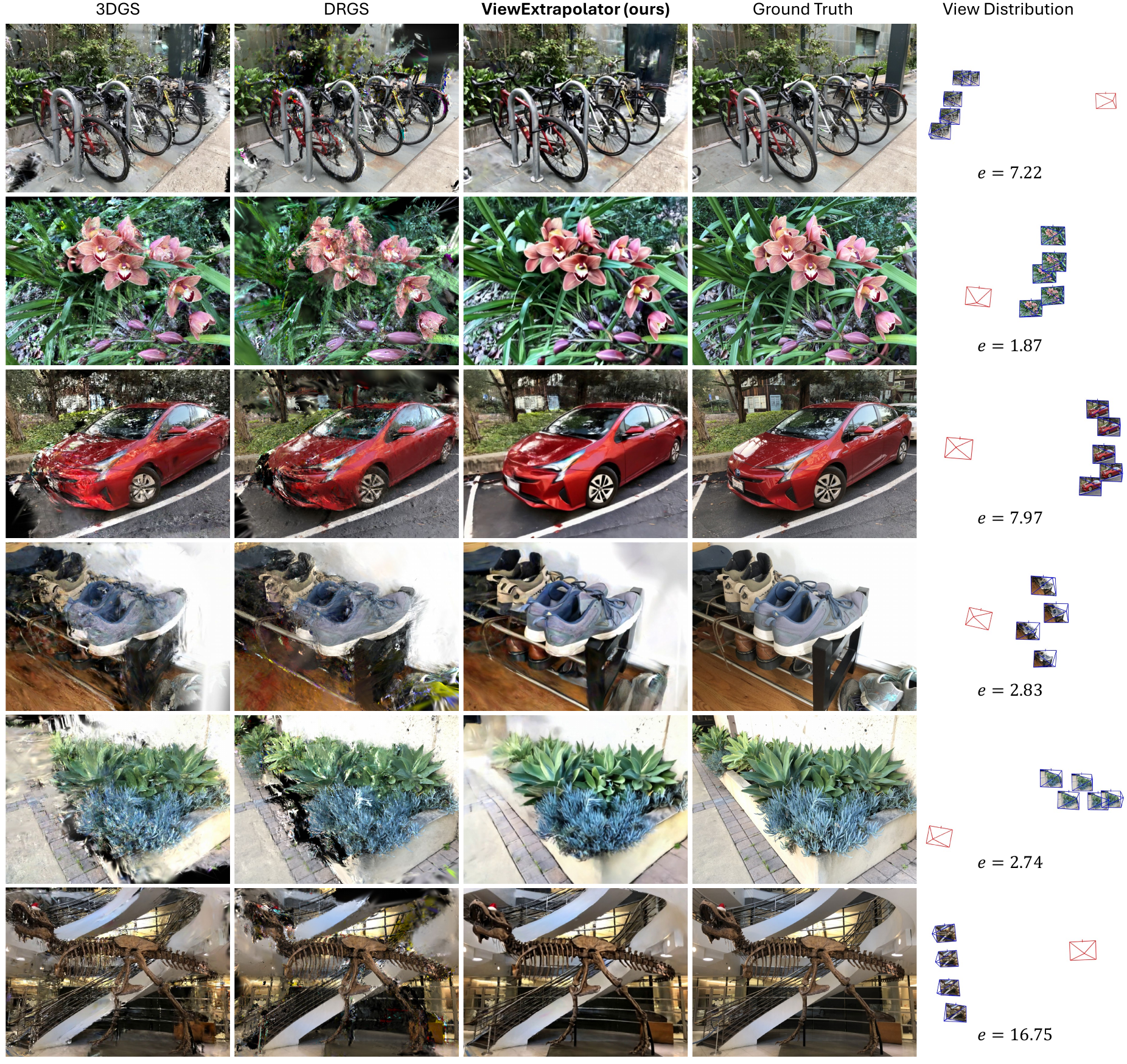}
        \captionof{figure}{\textbf{Additional comparisons.} We compare \name with 3DGS and DRGS on novel view extrapolation. \name demonstrates superior generation quality with much fewer artifacts. The last column shows the distribution of training and test views as well as the corresponding extrapolation degree $e$. Zoom in for details.
        }
        \label{fig:comparison2}
    \end{center}
}]

\begin{figure*}
    \centering
    \includegraphics[width=\linewidth]{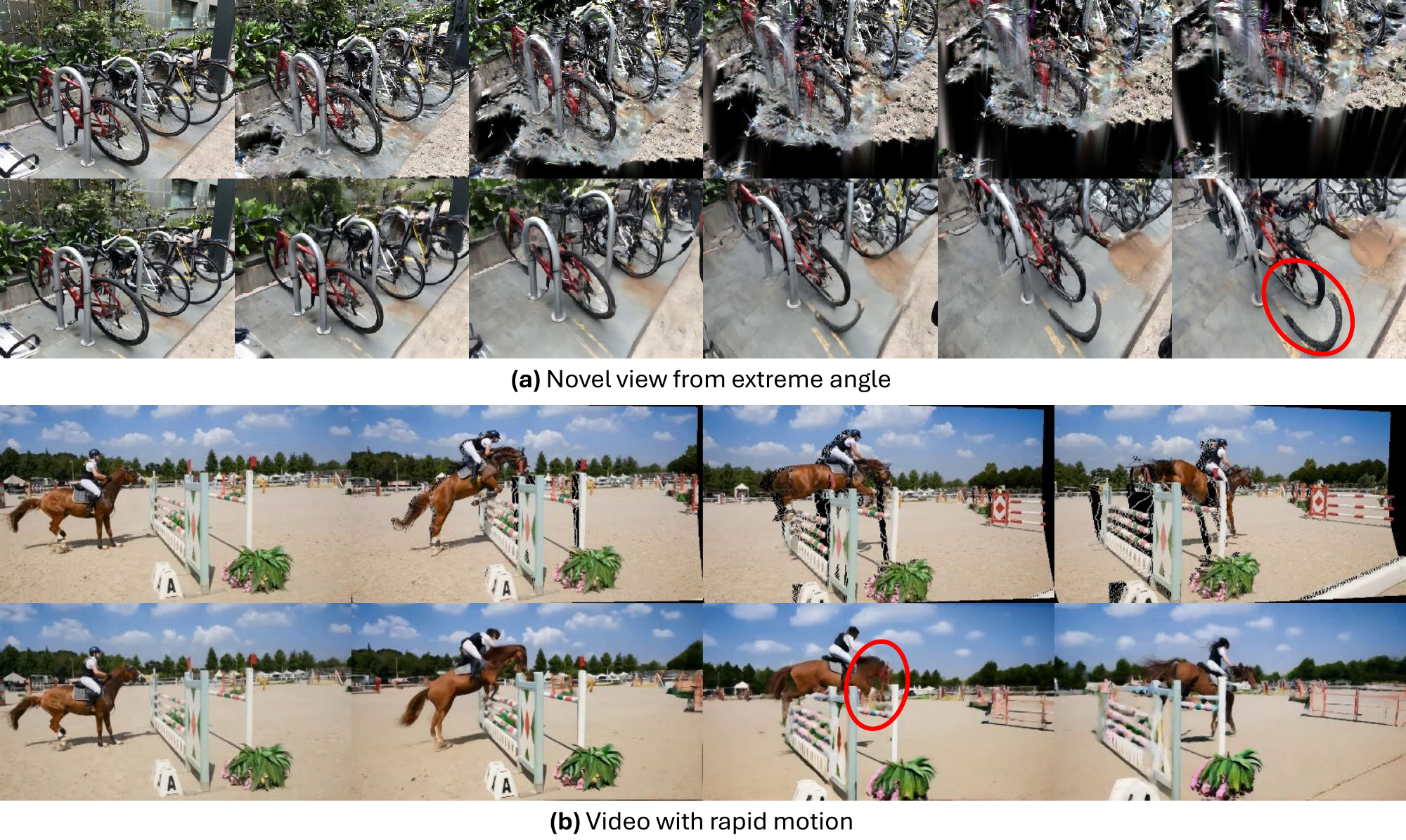}
    \caption{\textbf{Limitations and failure cases.} The generation quality would degrade when handling \textbf{(a)} novel views at extreme angles or \textbf{(b)} dynamic videos with rapid motion. (The top row in each section is the rendered artifact-prone video and the bottom row is the refined video.)}
    \label{fig:limitation}
\end{figure*}

\section{Additional Results}

We show additional qualitative comparisons in \cref{fig:comparison2}. Please visit the project page for video results: \url{https://kunhao-liu.github.io/ViewExtrapolator/}.

\section{Implementation Details}

\paragraph{Hyperparameters.}
We base our approach on the \texttt{xt-1-1} version of the SVD model, which generates 25-frame 6-fps videos at a resolution of $576 \times 1024$. For all experiments, we set $T=25$, $R=3$, and $R^\mathrm{guide}=1$, with $T^\mathrm{guide}=15$ for static scenes and $T^\mathrm{guide}=16$ for dynamic scenes. We set \texttt{noise\_aug\_strength }$=0$ to preserve the original scene content and set other parameters as default.
Our experiments were conducted on an NVIDIA RTX A5000 GPU with 24G memory, with each video refinement taking 3 minutes and 20 seconds.

\paragraph{Details on 3DGS Refinement.}
For the evaluations of novel view extrapolation, we employ the refined video frames (Ours (video) in \cref{tab:comparison}) to tune the pre-trained 3DGS model and evaluate renderings from the tuned 3DGS model (Ours (3DGS) in \cref{tab:comparison}). Given the refined video frames, we use them as well as the original training views to refine the 3DGS model using the standard L1, SSIM loss, and default densification strategy. In order to let the refined video frames regularize the geometry of 3DGS instead of being fitted as the view-dependent color, we incrementally increase the order of the spherical harmonics during refinement, starting from 0. In addition, to make the refined 3DGS more faithful to the original training views, we gradually decrease the frequency of training iterations that use the refined video frames throughout the training process. The refinement process requires one-third of the iterations used in the original 3DGS training.

\section{Limitations}
Although \name offers advantages in novel view extrapolation, it has several limitations. First, as an inference-stage approach, the quality ceiling of our method is bound by the original SVD model, meaning it also inherits certain drawbacks, such as lower resolution and color shifts. We believe incorporating more advanced video diffusion models could help enhance the overall quality. Second, our method encounters challenges when handling dynamic videos with rapid motion or extreme views where the novel views have very little overlap with the observed scene. We show the limitations and failure cases in \cref{fig:limitation}.

\end{document}